\documentclass[11pt,a4paper]{article}

\usepackage[utf8]{inputenc}
\usepackage[T1]{fontenc}
\usepackage{lmodern}
\usepackage{amsmath,amssymb}
\usepackage{microtype}
\usepackage{xcolor}
\usepackage[hyphens]{url}
\usepackage{hyperref}
\usepackage[margin=1in]{geometry}
\usepackage{enumitem}
\usepackage{listings}
\usepackage{tikz}
\usetikzlibrary{arrows.meta,positioning}

\hypersetup{
  pdfauthor={Luk Koska, Tom Koska, Ben Koska},
  colorlinks=true,
  linkcolor=blue!70!black,
  citecolor=blue!70!black,
  urlcolor=blue!70!black,
}

\title{Formalizing PTX in Lean to Prove GPU Kernel Correctness}
\author{Luk Koska \qquad Tom Koska \qquad Ben Koska\\[0.4em]
  \normalsize SF Tensor}
\date{}

\begin{document}
\maketitle

\begin{abstract}
We present an approach to mathematically formalizing PTX, an approach that allows modeling and reasoning
 about patterns common in GPU kernels for Deep Learning, including acquire-release, tensor cores, proxies,
async operations, multi-kernel pipelines and data-dependent control flow. We implement it in Lean and use it to prove the correctness of simple implementations of
a WGMMA GEMM kernel, FlashAttention, and a 3-kernel MoE pipeline.

\end{abstract}

\section{Introduction}
The correctness of GPU kernels is becoming increasingly critical in the age of AI.
Furthermore, as more and more kernels are written by LLMs, it is desirable to reduce human review burden, letting them trust outputs without extensive manual review.

A long line of work aims to verify the functional correctness of specific
classes of GPU kernels. For example, Volta~\cite{dubey2025equivalence} checks
equivalence of GPU kernels, supporting \texttt{\_\_syncthreads()}, \texttt{\_\_syncwarp()} and
synchronous tensor-core operations, but requires statically determined control
flow and excludes certain aspects of the PTX memory model.

Instead of focusing on specific classes of kernels, we mathematically model the
possible executions of a given PTX kernel in Lean, an approach that allows reasoning about many patterns used in modern GPU kernels,
including acquire-release synchronization and data-dependent control flow.
This approach lets LLMs write correctness proofs in a verifiable manner: only the mathematical specification of what the kernel should compute, the preconditions, our formalization and parsing of PTX in Lean and the Lean kernel, but not the LLM's outputs, need to be trusted.
Furthermore, it allows formalizing and checking human intuition for why a given kernel is correct.

\section{Related Work}\label{sec:related}

The axiomatic PTX memory model was formalized by Lustig et
al.~\cite{lustig2019formal} and extended with mixed-proxy
semantics~\cite{lustig2022mixed}; our model builds on both.
Dartagnan~\cite{tong2024unified} integrates the PTX and Vulkan models into a
bounded model checker for GPU code.
GPUMC~\cite{chakraborty2025gpumc} provides stateless model checking of GPU
programs under scoped-RC11, detecting races, barrier divergence, and
assertion violations.

Several tools verify properties of GPU kernels under the synchronous
CUDA/OpenCL programming model: GPUVerify~\cite{betts2012gpuverify} and
Faial~\cite{cogumbreiro2021faial} check race freedom (GPUVerify also
barrier divergence); PUG~\cite{li2010pug}, GKLEE~\cite{li2012gklee}, and
ESBMC-GPU~\cite{pereira2016esbmc} find races and other bugs by SMT solving,
concolic execution, and bounded model checking; test
amplification~\cite{leung2012amplification} generalizes single test runs
to classes of inputs; VerCors~\cite{blom2014gpgpu} proves functional
properties with permission-based separation logic.
Kojima and Igarashi~\cite{kojima2017hoare} give a Hoare logic for GPU
kernels, proving soundness and relative completeness for race-free programs
by relating lockstep and interleaving semantics.

Volta~\cite{dubey2025equivalence} checks
equivalence of ML GPU kernels and is complete for a well-defined class, but
requires statically determined control flow and memory-access structure.
Our model aims at generality (asynchronous features, data-dependent control
flow, multi-kernel pipelines) at the cost of manual, LLM-assisted proofs in
Lean rather than an automated decision procedure. Verified scheduling
rewrites for tensor programs have been formalized in
Coq~\cite{liu2022verified}, but at the level of a high-level language rather
than PTX.

LLM-assisted verification of generated CUDA kernels is also pursued by
ProofWright~\cite{chatterjee2025proofwright}, which combines VerCors safety
checking with Rocq-based semantic-equivalence proofs for a class of
elementwise kernels. Browne, Lee, and Salvi~\cite{browne2026proving} use a
deterministic translation of a CUDA subset into a sequential Lean
representation, Lean/Velvet for equivalence over \texttt{Float32} with opaque
transcendental functions, and VerCors for safety.

\section{Model}\label{sec:model}
Following Volta~\cite{dubey2025equivalence}, we model PTX rather than a higher-level language like CUDA, as most DSLs go through PTX in their lowering to SASS for NVIDIA GPUs.

\subsection{The PTX memory model}
The PTX memory consistency model, formalized by Lustig et
al., precisely specifies which values
loads may return through a system of ordering relations and axioms. This
enables formal reasoning about the possible outcomes of PTX programs, but
takes an after-the-fact view: it is built on the set of events that were
executed by each thread, without modeling control flow, making it difficult to
use directly for proving real kernels correct.

We take this as the basis of our model, and recall its most important aspects here.

Each thread's memory operations are ordered by \emph{program order}
($\mathrm{po}$), capturing that thread's sequential execution. A
\emph{reads-from} pair $(W,R)\in\mathrm{rf}$ records that read~$R$ obtains at
least one byte from write~$W$.
Each address has a write $W_0$ with its initial value. Synchronization between
threads is captured by \emph{synchronizes-with} ($\mathrm{sw}$); for example,
each thread's arrival at a \lstinline{bar.sync} barrier synchronizes with
every other thread's return from that barrier in the same generation.
Program order and synchronizes-with combine transitively into \emph{base
causality order}. The causality axiom constrains $\mathrm{rf}$: if write~$W$
causally precedes read~$R$ on the same bytes, $R$ cannot read from any write
earlier than~$W$ in coherence order $\mathrm{co}$ (\textit{a partial order on
writes to each byte of memory}).

To illustrate these relations, consider a CTA with two threads. The predicates
\texttt{p0} and \texttt{p1} select thread~0 and thread~1 respectively. The
global \texttt{u32} cell \texttt{x} is initially~0:

\begin{lstlisting}
@p0  st.global.u32 [x], 1;  // W: only thread 0 writes
     bar.sync 0;            // both threads arrive and wait
@p1  ld.global.u32 r, [x];  // R: only thread 1 reads
\end{lstlisting}

Both threads execute the same unconditional \lstinline{bar.sync}. Thread~0's
arrival ($A_0$) synchronizes with thread~1's return ($A_1$), giving the chain
\[
  W \xrightarrow{\mathrm{po}} A_0
    \xrightarrow{\mathrm{sw}} A_1
    \xrightarrow{\mathrm{po}} R.
\]

With no other writes to~$x$, the causality axiom forces
$W\mathrel{\mathrm{rf}}R$, so $r=1$. The full PTX model extends this account
with proxy and observation-order rules~\cite[PTX ISA, \S\S8.9--8.10,
9.7.15.1]{lustig2019formal,lustig2022mixed,ptxisa}.

\subsection{Modeling Decisions}

\textbf{Soundness.}\quad
We aim to make this model \textbf{sound}; that is, where NVIDIA's docs are
ambiguous, we interpret the spec \textbf{conservatively}, allowing
\textbf{more rather than less} outcomes of the kernel execution. That is, we prefer being unable to
prove correctness of some kernels, rather than being able to wrongfully prove
the ``correctness'' of an incorrect kernel.\footnote{Let $A$ be the set of executions allowed by PTX, and $B$ the set of executions allowed by our model.
 If $A \subseteq B$, then $\forall x \in B\, P(x) \implies \forall x \in A\, P(x)$}

\textbf{Modeling floats.}\quad
We model all floats as real numbers, with exact real-number operations (e.g.,
$\mathrm{fma}(a,b,c) = a \cdot b + c$,
$\mathrm{ex2}(x) = \exp(\ln 2 \cdot x)$). This does not allow for reasoning
about quantization error or numerical stability, meaning that only semantic,
not numerical, correctness can be proven. This choice is motivated by the observation that functional correctness is (a) binary and independent of the use-case, and (b) a precondition for numerical correctness to be meaningful.
We leave the question of ``\emph{will these numerics cause inference accuracy to decrease, or training to diverge?}'' as future work.

\subsection{Definition}
Consider a PTX kernel $P$.
We parse it into a list of instructions $(s_1, \ldots, s_n)$.

Each thread~$t$ executes this
program, one step at a time, affected by other threads only by means of
specific communication constructs (loads, barriers, etc.).

To decouple it from other threads, let us give it an \emph{oracle}~$o$ that
provides all information it needs from the outside world:
\[
  o \colon \mathbb{N} \to \mathsf{Value}
\]
That is, whenever a thread performs an event requiring external information
(e.g., a load, warp shuffle, or matrix-collective
result), the oracle provides the information it needs. The $i$-th such event
gets the value $o(i)$.

Given this oracle, the thread~$t$ can be given deterministic small-step
semantics. A thread's state is a tuple $(pc, R, k)$, where $pc$ is the
program counter, $R$~is the register file (mapping register names to typed
values), and $k$~is the index of the next oracle query. One step is a
deterministic function
\[
  \mathsf{step} \colon \mathsf{State} \to
    \mathsf{State} \times \mathsf{Event}^*
\]
mapping the current state to the next state and zero or more emitted events
(reads, writes, barrier arrivals, etc.). A single instruction may consume
multiple oracle queries; for example, a vector load. Arithmetic on
register operands emits no events and makes no oracle queries. Branches read
a predicate register and set $pc$ to the target or to $pc + 1$; predicated instructions
are similarly skipped when the predicate is false. Since register values may
depend on oracle answers, the path a thread takes, including loop iteration
counts, can depend on what values it reads from memory.

Letting reads return arbitrary values and checking consistency afterwards dates back at least to Batty et al.~\cite{batty2011mathematizing}.

\medskip

Now, define a kernel launch as a tuple
\[
  K := (P, G, I)
\]
whereby $P$~is the program, $G$~is the launch grid (encoding grid, cluster,
and block dimensions, launch parameters, and allocation layout), and $I$~is a
total map from byte addresses to initial byte values, fixed for the launch.
Launch preconditions constrain initialized inputs; other bytes have arbitrary
but fixed initial values, as in the ISA's initialization rule~\cite[\S8.2.6]{ptxisa}.
Correctness quantifies over every such initial map.

To reason about what the kernel ``may do'', we want to define the set of all
allowed executions of this program. An execution of~$K$ assigns an oracle to
every launched thread and runs each thread for some finite number of
steps (\textit{not necessarily until it exits; executions in which every thread has
exited are called \emph{complete} below}) producing a combined set of memory
events. An execution is \emph{consistent} if a bytewise reads-from relation
$\mathrm{rf}$ (with initialization as a possible source), a per-byte
coherence order $\mathrm{co}$, an order $\mathrm{sc}$ on \lstinline{fence.sc}
operations, and the construct-specific matching (barrier generations,
mbarrier phases, async completion groups, etc.)\ can be chosen so that:

\begin{itemize}[nosep]
  \item \textbf{Value agreement.} Each read has a unique reads-from
    source per byte (initialization or a write) and receives its value.
    The exceptions are \emph{tainted} reads, which have weaker source and
    value constraints: a read is tainted if it participates in a uniform-size
    race (a morally weak conflict unordered by causality) or reads a byte on
    which two writes race. Furthermore, if the execution contains
    mixed-size data races~\cite[\S\S8.7.1--8.7.2]{ptxisa}, none of these guarantees apply.
  \item \textbf{Memory-model axioms.} Under the same no-mixed-size-race
    condition, the coherence, atomicity, SC-per-location, and causality
    axioms of the PTX memory consistency
    model~\cite{lustig2019formal,ptxisa} are satisfied.
  \item \textbf{Construct rules.} Barrier matching, mbarrier phase
    completion, async-operation completion, and other synchronization
    constructs satisfy their respective protocol requirements, as explained
    later in Section~\ref{sec:aspects}.
\end{itemize}

We omit the no-thin-air axiom (discussed in Section~\ref{sec:nta}) and the
Fence-SC axiom~\cite[\S8.10.2]{ptxisa}. Fences related by $\mathrm{sc}$ still
synchronize in that direction~\cite[\S8.9.4]{ptxisa}. Omitting these two axioms
weakens consistency, admitting more executions.

An execution is \emph{complete} when every launched thread
has exited. Correctness is stated under
launch preconditions on launch parameters, layout, alignment, input initialization
and nonaliasing. For every launch satisfying the stated preconditions, every
consistent, complete execution of our model satisfies the postcondition, and
at least one such execution exists. For each verified variant, we also
exhibit a concrete launch satisfying its preconditions.
We construct an execution for each admissible launch to rule out vacuity
(if there are no terminating executions, the statement
``all terminating executions satisfy the postcondition'' is vacuously true).
This existence proof is not needed for the logical validity of the
partial-correctness statement; rather, it provides an additional consistency check
(a mistake in the modeling constraints could otherwise exclude every complete
execution, making the correctness theorem vacuously true).
Applying these results to real PTX assumes that the kernel is free of
ISA-level undefined behavior and that \texttt{ptxas} and the hardware implement
PTX correctly. We do not prove the UB-freedom assumption
(Section~\ref{sec:ub}).

The model and all proofs are formalized in Lean~4; the memory-model axioms
are expressed as definitions and predicates on execution candidates.

\subsection{Final State of Global Memory}

A kernel's contract specifies the state of global memory after execution
completes. Since our model constructs an execution graph rather than
advancing a single mutable heap, the final memory needs an explicit
definition: what bytes would a synchronized observer read after the kernel
has finished?

We answer this by adding \emph{observer read events} to the execution
candidate, whose footprints together cover every allocated global byte.
These are logical events
representing an observation after synchronized kernel completion. API
synchronization~\cite[\S8.9.4]{ptxisa} orders every kernel event before these
observer reads: the end of all kernel threads synchronizes with the end of the
kernel, which synchronizes with the subsequent observation. These
synchronizes-with edges also participate in proxy-preserved base causality,
providing causality for overlapping bytes.

\begin{figure}[t]
\centering
\begin{tikzpicture}[
    node distance=2.8cm,
    every node/.style={font=\small},
    >=Stealth,
    solid edge/.style={->, thick},
    hyp edge/.style={->, thick, dashed},
    derived edge/.style={->, thick, densely dotted},
  ]
  \node (W0) {$W_0$: init $x{=}0$};
  \node (W)  [right=of W0] {$W$: store $x{=}1$};
  \node (R)  [right=of W]  {$R$: observer read};

  \draw[solid edge] (W0) -- node[above]{\scriptsize co} (W);
  \draw[solid edge] (W) -- node[above]{\scriptsize cause (API sync)} (R);
  \draw[hyp edge]   (W0) to[bend right=35]
    node[below]{\scriptsize rf\,? (hypothesized)} (R);
  \draw[derived edge] (R) to[bend right=35]
    node[above]{\scriptsize fr (derived)} (W);
\end{tikzpicture}
\caption{Observer cycle excluding a stale read.  Solid edges are fixed by the
execution; the dashed edge is the hypothesized stale source.  If $R$ reads
from initialization ($W_0 \xrightarrow{\mathrm{rf}} R$), then because $W_0$
is coherence-before $W$, the from-read edge $R \xrightarrow{\mathrm{fr}} W$
is derived.  This contradicts $W \xrightarrow{\mathrm{cause}} R$, so the
causality axiom excludes $W_0$ as a source.}
\label{fig:observer}
\end{figure}
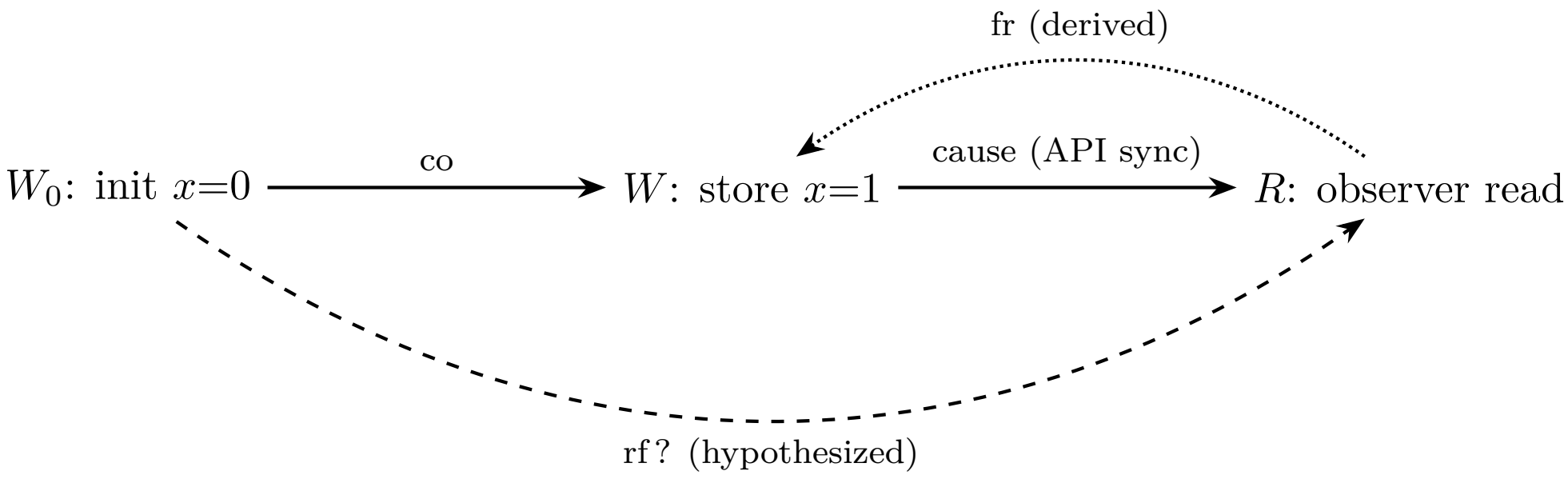

The observer values are constrained by the same memory-consistency rules as
any other read. For a simple example: if a single store writes $x = 1$ to a
cell initialized to $0$ and no other write to $x$ exists in the execution,
API synchronization
places the store before the observer read in causality. The causality axiom
then excludes the initial write as the reads-from source, and value agreement delivers
$x = 1$. Figure~\ref{fig:observer} visualizes why: if the observer were to
read from initialization instead, the \emph{from-read} edge $R
\xrightarrow{\mathrm{fr}} W$ (derived because $R$ would read from a write
coherence-before $W$~\cite[\S8.9.7]{ptxisa}) would contradict the causal
edge $W \xrightarrow{\mathrm{cause}} R$, violating the causality
axiom~\cite[\S8.10.6]{ptxisa}.

The observer payloads define one output byte map for each candidate
execution. Correctness then requires that this map satisfies the
postcondition in every admitted execution. This output map also serves as the
input for a subsequent kernel launch on the same stream, enabling
sequential composition (Section~\ref{sec:aspects}).

\subsection{Modeling Aspects of PTX}\label{sec:aspects}

The base definition handles ordinary loads, stores, arithmetic, and
data-dependent control flow: a load gets its value from the oracle and
records a read event; a store records a write event with the computed
address and value; branches follow the program deterministically given
the current register state. These ordinary accesses use the generic proxy
and are constrained by the consistency axioms without additional
extensions.

Real kernels also use synchronization and communication primitives, each
of which adds structure to the model. We explain how each is formalized.

\textbf{CTA barriers.}\quad
A CTA barrier (\lstinline{bar.sync}) is modeled with separate arrival and
return events. Each use of a barrier resource is assigned a natural-number
\emph{generation identifier}, grouping arrivals and returns for that use; the
same resource can be reused for later generations. In the full-CTA case, a
return in generation~$g$ requires that every CTA thread has an arrival in
generation~$g$. A thread that has exited is exempt, provided all of its
arrivals on that barrier are in generations earlier than~$g$.
Each arrival in generation~$g$
synchronizes with each return in the same generation, adding an
$(\mathrm{arrival}, \mathrm{return})$ pair to the $\mathrm{sw}$ relation.
This mechanism is used in the introductory example, where a write before
the arrival becomes causally visible to a read after the
return~\cite[\S\S9.7.15.1, 8.9.4]{ptxisa}.

\textbf{Proxies.}\quad
A proxy describes the \emph{method of accessing memory}~\cite[\S\S8.5--8.6]{ptxisa}. In the model, every
memory event carries a proxy tag. Ordinary loads and stores use the
\texttt{generic} proxy; asynchronous bulk copies use the \texttt{async}
proxy; TMA descriptor reads use \texttt{tensormap}. Two accesses to the
same bytes through different proxies are not ordered by an ordinary
base-causality path alone~\cite{lustig2022mixed}. To establish ordering across
proxies, the model requires a \emph{proxy fence}~\cite[\S9.7.15.4]{ptxisa}:
an explicit fence event whose direction and byte range match the endpoints,
positioned on the base-causality path between them. Certain
instruction-specific completion rules also supply the needed cross-proxy
visibility~\cite[\S\S9.7.10.28.2, 9.7.17.4]{ptxisa}; for example, a
successful acquiring mbarrier wait that observes a TMA copy's completion
can make its async-proxy data writes visible to a later generic-proxy
read~\cite[\S\S9.7.10.28.2, 9.7.15.16.19]{ptxisa}.

\textbf{Asynchronous copies.}\quad
PTX asynchronous copy instructions (\lstinline{cp.async},
\lstinline{cp.async.bulk}) execute their data transfers outside the issuing
thread's program order~\cite[\S8.9.1.1]{ptxisa}. We model this by giving
each issued copy a distinct logical identity: its source reads and
destination writes are events in the same execution candidate as ordinary
thread events, but they are not ordered by the issuer's $\mathrm{po}$. The
issuing thread's issue event is recorded as an ordinary event of that thread.
The copy is ordered after the issuer's earlier instructions, and its writes
are guaranteed visible only to events causally after a wait that completes
it, e.g., \lstinline{cp.async.wait_group}~\cite[\S\S8.9.1.1, 9.7.10.28.1]{ptxisa}.

\textbf{WGMMA.}\quad
The register results of a warp-group matrix
multiply-accumulate~\cite[\S\S9.7.17.5, 9.7.17.7]{ptxisa} are oracle values.
The model constrains them to the matrix product plus accumulator only when
the operation is synchronized as the ISA requires; with insufficient
synchronization the results remain unconstrained, so a proof cannot rely on
them.

\textbf{Warp shuffles.}\quad
A warp shuffle (\lstinline{shfl.sync}~\cite[\S9.7.10.6]{ptxisa}) lets lanes
within a warp exchange register values. Each shuffle event records the lane's
input value and receives an oracle result. Shuffle instances across
participating lanes are grouped into a matched set by a natural-number
generation, analogous to barrier generations. Within a matched group, each
lane's oracle output is constrained to equal the appropriate partner lane's
recorded input, as determined by the shuffle mode.

\textbf{Cluster Launch Control.}\quad
Our model requires that all launched threads must exit. A
successful cancellation atomically removes a not-yet-running cluster, whose
threads have empty histories and thus never reach an exit. We handle this
by exempting canceled threads from the exit requirement. Distinct
successful cancellation operations in the same grid cannot identify the
same canceled cluster~\cite[\S9.7.15.18]{ptxisa}.
This extension has not been formalized in Lean.

\textbf{Multi-kernel pipelines.}\quad
One step of model inference or training generally consists of several kernel
launches. We want to be able to prove the end-to-end correctness of such a
full step. To this end, we formally model the result of launching several
kernels sequentially on the same CUDA stream. API synchronization orders the
completion of one kernel before the start of the
next~\cite[\S8.9.4]{ptxisa}. At completion, the kernel's writes are
flushed~\cite[\S4.15.1]{cudaguide}, and the driver invalidates global L1
cache lines between dependent grids so that the second kernel correctly
fetches the first kernel's stores~\cite[\S9.7.10.1]{ptxisa}. We model this
boundary by equating the successor kernel's initial global byte map with the
predecessor's output map (with the exception of the tensormap
proxy~\cite[\S8.9.4]{ptxisa}).

\section{Applications}\label{sec:applications}

To illustrate the expressiveness and usability of the model, we have used it to prove the correctness of several GPU kernels, as outlined below.
For every admissible launch of each variant, we prove that all consistent,
complete executions of our model produce the specified output, and separately
construct such an execution to establish non-vacuity.

The considered PTX code is \textbf{hand-written} rather than real compiler output (which commonly contains inlining and arithmetic optimizations, making proofs more difficult).
All kernels were tested for correctness against a reference on an H100 and B200 (except for the WGMMA kernel, which was only tested on the H100).

\begin{itemize}[nosep]
  \item \textbf{Elementwise.} An affine map $y_i = \alpha x_i + \beta$ in f32
    over a vector of length~$N$. A 1-D grid assigns one thread per element;
    out-of-range threads exit.
    (18~instructions; 1{,}077~proof lines.)

  \item \textbf{Scalar matmul.} $C = AB$ with
    $A \in \mathbb{R}^{M\times K}$, $B \in \mathbb{R}^{K\times N}$, all f32
    row-major. A 2-D grid of 2-D CTAs assigns one thread per output element;
    each thread loops an \lstinline{fma} over $k < K$ ($K = 0$ is allowed and
    gives zero). The proof uses an accumulation invariant over the
    inner-dimension loop.
    (42~instructions; 1{,}923~proof lines.)

  \item \textbf{Ampere tiled matmul.} f16 inputs ($A$ row-major $M\times K$,
    $B$ column-major $K\times N$), f32 output. One warp computes one
    $16\times 8$ output tile, looping over~$K$ in steps of~16:
    \lstinline{cp.async} copies the current $A$ and $B$ tiles into shared
    memory, \lstinline{cp.async.wait_group} and \lstinline{bar.sync} publish
    them, and \lstinline{mma.sync.aligned.m16n8k16} accumulates; the same
    shared-memory buffers are reused in every iteration. Requires $16\mid M$,
    $8\mid N$, $16\mid K$.
    (94~instructions; 5{,}418~proof lines.)

  \item \textbf{Hopper matmul.} $C = AB$ with f16 inputs, f32 output,
    $K = 16T$, $T \ge 2$. One warpgroup computes one $64\times 8$ output
    tile, looping over the $T$ slices through a two-stage shared-memory
    pipeline whose buffers are refilled by TMA
    (\lstinline{cp.async.bulk.tensor.2d}), tracked by mbarriers. At each
    step every thread issues one \lstinline{wgmma.mma_async.m64n8k16} that
    accumulates into the same registers.
    Successful acquiring mbarrier waits ensure that the TMA copies have
    completed before WGMMA reads their operands.
    \lstinline{wgmma.fence} orders prior accumulator-register accesses before
    WGMMA, while \lstinline{wgmma.commit_group} and
    \lstinline{wgmma.wait_group 0} ensure completion before the accumulated
    results are read.
    (98~instructions; 16{,}973~proof lines.)

  \item \textbf{FlashAttention}~\cite{dao2022flashattention}.
    Computes
    $O = \mathrm{softmax}(\alpha\, Q K^{\top})\, V$ in exact real arithmetic,
    $Q \in \mathbb{R}^{M\times D}$, $K, V \in \mathbb{R}^{N\times D}$.
    The scalar kernel assigns one thread per output element
    (67~instructions; 2{,}045~proof lines).
    The proof uses the online softmax invariant.

  \item \textbf{Mixture of Experts.} An MoE layer with 2 experts and top-1 selection gets $N$ input tokens,
   handling routing, expert computation (simple linear layers) and combine.
    (120~instructions across the three launches; 5{,}722~proof lines.)

  \item \textbf{Simple megakernel.} Two CTAs use atomic adds to a global work
    counter (as is commonly used in megakernels to distribute work) and
    data-dependent control flow to execute the job received by each thread. We
    describe our proof of this kernel's correctness below.
    (15~instructions; 876~proof lines.)
\end{itemize}

\subsection{Proof sizes and checking}\label{sec:checking}

In total, the kernels described above comprise 9 PTX entry points and
454~instructions, verified by 34{,}034 lines of proof\footnote{Proof lines
are nonblank, noncomment lines of \texttt{theorem} declarations (statements
and proofs), classified by Lean's parser. The kernel proofs also build on
about 5{,}600 lines of shared lemmas not counted per kernel, and the Hopper
kernel reuses about 2{,}100 lines of lemmas from a simpler single-stage
variant not described here.}. In contrast, our model of Section~\ref{sec:model} is made up of 2{,}629 lines of
definitions.

\textbf{Checking.}\quad All proofs are checked by Lean~4 (v4.35.0) with
Mathlib. The development contains no \textit{sorry} and no axiom declarations
(the memory-model axioms are ordinary definitions), and every theorem is verified by
\texttt{\#print axioms} to use only Lean's three standard axioms. Every variant
has both a universal correctness theorem and a constructive non-vacuity
theorem, instantiated at a concrete launch.

\subsection{Correctness proof of a simple megakernel}
\noindent\begin{minipage}{\linewidth}
We illustrate a simple application of the model by summarizing a correctness proof of a simplified megakernel.

The kernel launches two CTAs of one thread each.  Three disjoint global \texttt{u32} cells are given: a counter initialized
to~$0$ and outputs $x$, $y$.  With \lstinline{
\lstinline{

\begin{lstlisting}
atom.global.relaxed.gpu.add.u32 %r0, [%rd0], 1;
setp.eq.u32 %p0, %r0, 0;
@%p0 bra ITEM_A;
setp.eq.u32 %p1, %r0, 1;
@%p1 bra ITEM_B;
bra DONE;
ITEM_A:
    mov.u32 %r1, 1;
    st.global.u32 [%rd1], %r1;
    bra DONE;
ITEM_B:
    mov.u32 %r2, 2;
    st.global.u32 [%rd2], %r2;
DONE:
    ret;
\end{lstlisting}
\end{minipage}

We prove that every consistent, complete execution of our model yields
$(x, y, \mathit{counter}) = (1, 2, 2)$.

\medskip
\textbf{Ticket uniqueness.}\quad
In our oracle model, the atomic return values begin unconstrained.  In a
complete execution both threads have reached their exit, so each has
performed exactly one increment; these are the counter's only modifications.
All kernel accesses cover one of the three disjoint \texttt{u32} cells;
together with the API-synchronized observer, this rules out mixed-size
races and makes the memory-model axioms applicable.  The GPU-scoped relaxed
increments are mutually morally strong, so their counter reads are
untainted.  Coherence orders the two counter writes.  SC-per-location
excludes reading an increment's own write or a later write in that order.
RMW atomicity prevents skipping an earlier modification: the first
increment in coherence order therefore reads initialization on every byte,
and the second reads the first increment's write.  Value agreement gives
returns $0$ and~$1$ and writes $1$
and~$2$~\cite[\S\S8.7--8.7.2, 8.9.6, 8.10.3, 8.10.5]{ptxisa}.  The
returned tickets are $\{0,1\}$ in some order; no CTA is assigned a
predetermined job, and no inter-CTA causality between the relaxed atomics
is required.

\textbf{Deterministic dispatch.}\quad
Since the tickets are $\{0,1\}$, exactly one thread enters each branch.  The
thread receiving~$0$ stores $x = 1$; the thread receiving~$1$ stores
$y = 2$.  Each output cell therefore has exactly one writer, and the two
output writes are to disjoint cells.

\textbf{Final observation.}\quad
After kernel completion, a synchronized observer reads every allocated global
byte.  API synchronization orders each kernel write before the observer's
corresponding read.  For the bytes of these three cells, the causality axiom
excludes initialization and earlier modifications as stale sources, and
value agreement delivers the coherence-latest
write~\cite[\S\S8.9.4, 8.10.6]{ptxisa}.  For $x$ and $y$, the unique kernel write to each
cell is coherence-latest; for the counter, the coherence-last increment
wrote~$2$.  No causal order between the two counter writers is required.  The
observer therefore sees $(x, y, \mathit{counter}) = (1, 2, 2)$.

\section{Limitations and Future Work}
\subsection{Instruction coverage and larger kernels}
The interpreter implements 81 instruction forms (66 scalar, control-flow,
and memory instructions; 5 Ampere tensor-core and async-copy forms; 10
Hopper forms), each with fixed qualifier combinations, primarily those needed by
the kernels of Section~\ref{sec:applications}. Using the model for a wide
variety of production kernels will require broader coverage, including
formalization of tcgen05 for Blackwell-era kernels and cluster-level features
such as distributed shared memory.

Furthermore, the model encodes floating-point values symbolically as
fragments of a real number rather than as IEEE bits. Bit-level
reinterpretation of floats is therefore outside the model (integer
manipulation of exponent and mantissa bits for fast exponential
approximations, packing formats by shifting bits, and round
trips like \lstinline{ld.f32} followed by \lstinline{st.b32}).

The kernels we reasoned about were hand-written to keep them simple;
optimized compiler outputs commonly contain inlining and non-obvious
arithmetic optimizations (e.g., eliminating integer divisions), which may
complicate proofs. More fundamentally, the proofs average about 75 lines of
proof per PTX instruction (34{,}034 proof lines for 454 instructions; the
pipelined Hopper kernel alone accounts for about 17{,}000 lines). Verifying
substantially larger kernels will likely need more proof automation and
reusable invariants.

\subsection{Proof of Termination}

Certain synchronization constructs, like polling a flag for acquire-release
patterns, are represented as unbounded loops in PTX. For example, an
acquire-release handshake may compile to a loop that repeatedly loads a flag
until a release store from another thread becomes visible. Our model admits
arbitrary finite loop lengths but does not prove that such loops terminate.

Alternatively, one could manually pattern-match such constructs, replacing
them with a special ``wait until this event occurs'' instruction whose progress
guarantees would need to be justified.

Our results are therefore partial-correctness guarantees: they rule out wrong
answers on completed runs but do not prove termination. In practice, runtime
timeouts and checkpoint-based restarts can support automatic detection and
recovery from kernel hangs, at the cost of restart overhead and repeating
work since the last checkpoint. A silent functional-correctness bug, by
contrast, can propagate undetected through an entire training run, with far
greater economic impact.

\subsection{Handling Undefined Behavior}\label{sec:ub}

Applying our model-level correctness results to real PTX assumes that the
kernel is free of ISA-level undefined behavior under its launch preconditions,
and that \texttt{ptxas} and the hardware implement PTX correctly. We do not
establish that external UB-freedom assumption.
Developing techniques for proving that a kernel cannot encounter UB is an
interesting direction for future work, but it is not straightforward. We do
not have a precise characterization of which UB-based \texttt{ptxas}
transformations are permitted.

\subsubsection*{Why proving UB-freedom is hard}

The difficulty is that whether a thread reaches an undefined operation can
depend on values it reads from memory, which in turn depend on what other
threads have written, and those writes may themselves be reordered by the
compiler. A simple attempt at a criterion and two complications illustrate
the problem.

\begin{samepage}
\textbf{Checking prefixes.}\quad
A natural approach is to check whether any consistent partial execution
reaches an undefined operation. Consider two threads with global u32
locations $x=y=0$, using \lstinline{.relaxed.gpu}
loads and stores so that overlapping accesses are morally strong:

\begin{lstlisting}
T0 (Thread 0)              T1 (Thread 1)
r1 = load x                r2 = load y
if r1 == 1: UB             store x, r2
store y, 1
\end{lstlisting}
\end{samepage}

Here \texttt{UB} could, for example, be a misaligned u32 access to a separate
allocation~\cite[\S6.4.1]{ptxisa}. Stopping T0 just before that misaligned operation
removes its store to $y$. T1 can then read only $y=0$, so it cannot supply
$x=1$. No consistent partial execution of this form reaches UB.

If, however, the compiler moves T0's constant store to $y$ before its load of $x$, the following can happen:
T0 can write $y=1$, T1 can read it and write $x=1$, and T0 can then
read $x=1$ and encounter UB. This schedule is \textbf{not} precluded by the No-Thin-Air axiom.

The issue here is that, because the compiler can reorder operations (subject to the memory model's constraints), instructions coming after
the first UB in PTX program order might actually be executed \textbf{earlier} by the hardware, requiring that one take the impact of later
operations into account when reasoning about UB.
Furthermore, one needs constraints for what \textbf{can't} execute before the UB; if the instruction after UB is impacted by it, but could
also be ``reordered'' to execute before it, the UB might ``justify itself''.

\subsection{No Thin Air}\label{sec:nta}

The formal PTX memory model~\cite{lustig2019formal} states the no-thin-air
axiom as
\[
  \operatorname{acyclic}(\mathrm{rf} \cup \mathrm{dep}),
\]
where $\mathrm{dep}$ is a dependency relation. Here, $\mathrm{dep}$ denotes
\textbf{semantic} rather than syntactic dependencies, to allow the compiler to
perform optimizations that break dependency
chains~\cite[\S8.10.4]{ptxisa}:

\begin{quote}
the implementation may perform any chain of reasoning to determine that a
store is not dependent on a prior load, and thus break self-fulfilling cycles
which would otherwise apparently be forbidden by the No-Thin-Air axiom.
\end{quote}

A syntactic register-use chain is not sufficient. Consider:

\begin{lstlisting}
ld r1, [x]
mul r2, r1, 0
st [y], r2
\end{lstlisting}

Although \texttt{r2} syntactically depends on \texttt{r1}, the store's value
is semantically independent of the load: the integer multiply-by-zero always
produces zero regardless of the loaded value.

Using the no-thin-air (NTA) axiom soundly would therefore require proving that a
semantic dependency exists, not merely that a syntactic register chain
connects two instructions. We thus leave NTA out of our formalization. This expands the set of allowed executions, so the universal output proof must cover at
least as many executions. Our kernel proofs instead use coherence, atomicity,
and synchronization to constrain read values.

No-thin-air has been an active area of research in the C++ memory
model~\cite{batty2011mathematizing,batty2015problem,kang2017promising}, and
the new Tile IR memory model~\cite[\S8.9]{tileirmodel} explicitly forgoes a
formal no-thin-air axiom, retaining only an informal assurance that the
implementation will not provide values out of thin air.

\subsection{Modeling SASS}

One route around the PTX compiler and the ambiguity around the $\mathrm{dep}$
relation is to model a fixed SASS binary instead of PTX. Compiler transformations
are already reflected in its instruction stream. The hardware issues SASS
instructions in order and does not perform branch prediction or speculative
execution~\cite[\S3.2.2]{cudaguide}, although memory accesses may still complete
or become visible out of order. This gives the following directions:

\textbf{No-Thin-Air.}\quad
In PTX, the No-Thin-Air axiom relies on semantic dependencies, making it difficult to use as described in Section~\ref{sec:nta}.
In SASS, with the possibility of the compiler breaking dependencies through optimizations eliminated, this rule could be reduced to considering syntactic dependencies.

\textbf{Proving freedom of UB.}\quad
If a SASS execution encounters UB, consider the prefix ending just before
its first occurrence. That prefix is UB-free, so we can analyze it using
the normal execution rules. The approach is to show that no such prefix
can take a step that causes UB. These partial executions include any
memory or asynchronous operations still in flight.

We note that SASS, at the time of writing, is not fully documented by NVIDIA~\cite{baddouh2027eoss}.

\subsection{Concurrent, Multi-GPU, and Multi-Node Kernels}

The model supports sequential kernel launches on one GPU. It does not yet
cover concurrent modification of memory by the host, concurrently executing
kernels, or other GPUs in the case of multi-GPU kernels, which are common for collectives and expert parallelism.
Extending to multi-node communication would require
modeling the transport layer's ordering and completion rules together with
their interaction with GPU memory visibility.

\subsection{Non-Coherent Loads}

PTX's memory consistency model excludes texture accesses, including
\lstinline{ld.global.nc}~\cite[\S8.1]{ptxisa}. These loads use a non-coherent
read-only cache~\cite[\S9.7.10.9]{ptxisa}. Extending the model to them requires
a separate account of their visibility guarantees and read-only requirements;
this is future work.

\subsection{Integration with TorchLean}

TorchLean~\cite{george2026torchlean} can import captured PyTorch computation
graphs into Lean and give them a formal semantics; using these as the
specification of what a kernel should compute could alleviate the need for
humans to review a hand-written Lean specification.

\section*{Outlook}

We have successfully used the presented model to reason about many common patterns in modern kernels
(async tensor core ops, multi-kernel pipelines, dynamic work
distribution via atomics and data-dependent control flow). A model of SASS, or a more rigorous definition of undefined
behavior at PTX level, could make undefined behavior far more tractable to
treat.

We believe in a future where the entire software stack of frontier training
runs is formally proven to be correct, supported by the increasing efficiency
and capability of LLMs in constructing Lean proofs, along with a library of useful lemmas to
reduce the per-kernel proof effort toward the novel ideas it contains.

\section*{AI Usage Statement}
Codex and OpenCode with GPT 6 Astra, Gemini 4 Argon, Opus 5.5 and Fable 5.1 assisted in implementation of the Lean model and the kernel correctness proofs.
The model, PTX kernels and the specification of what they compute were carefully reviewed by us; the proofs contained no custom axioms or sorrys and were checked by the Lean kernel.

Parts of the paper were written by Opus 4.6. All generated content was carefully reviewed by us.

\smallskip
{%
\let\oldthebibliography\thebibliography
\renewcommand{\thebibliography}[1]{%
  \oldthebibliography{#1}%
  \setlength{\itemsep}{0pt plus 1pt}%
  \setlength{\parsep}{0pt}%
  \setlength{\parskip}{0pt}%
  \footnotesize
}
\bibliographystyle{plain}
\bibliography{refs}
}

\end{document}